\pdfoutput=1
\documentclass[11pt,xcolor=table,xcdraw]{article}

\usepackage[]{ACL2023}

\usepackage{times}
\usepackage{latexsym}
\usepackage{enumitem}
\usepackage{makecell}
\usepackage{multirow}
\usepackage{subfigure}
\usepackage[T1]{fontenc}
\usepackage[utf8]{inputenc}

\usepackage{microtype}

\usepackage{inconsolata}

\usepackage{todonotes}
\usepackage{booktabs}

\usepackage{float}

\usetikzlibrary{shadows,calc}
\usepackage{tcolorbox}
\tcbuselibrary{skins,theorems,breakable}

\tcbset{
mybox/.style={
  breakable,
  enhanced,
  boxrule=0.4pt,titlerule=-0.2pt,drop fuzzy shadow,
  width=\linewidth,
  fonttitle=\Large\bfseries\sffamily,
  fontupper=\sffamily,
  fontlower=\sffamily,
  before=\par\medskip\noindent,
  after=\par\medskip,
  center title,
  toptitle=3pt,
  top=11pt,topsep at break=-5pt,
  frame style={left color=red!75!black,
right color=blue!75!black},
  colback=white
}}

\newfloat{textbox}{thp}{lop}
\floatname{textbox}{Box}

\title{Responsibility Perspective Transfer for Italian Femicide News}

\author{Gosse Minnema$^{a\ast}$, %\hspace{2em} 
Huiyuan Lai$^a$\thanks{\ \ Shared first co-authorship.}\ , %\\ 
\textbf{Benedetta Muscato}$^b$ and %\hspace{2em} 
\textbf{Malvina Nissim$^a$} \AND \textnormal{$^a$University of Groningen, The Netherlands}\\$^b$University of Catania, Italy\\\texttt{\{g.f.minnema,h.lai,m.nissim\}@rug.nl}}

\begin{document}
\maketitle

% word limit for abstract on SoftConf: 300 w

\begin{abstract}
Different ways of linguistically expressing the same real-world event can lead to different perceptions of what happened. Previous work has shown that different descriptions of gender-based violence (GBV) influence the reader's perception of who is to blame for the violence, possibly reinforcing stereotypes which see the victim as partly responsible, too. As a contribution to raise awareness on perspective-based writing, and to facilitate access to alternative perspectives, we introduce the novel task of automatically rewriting GBV descriptions as a means to alter the perceived level of responsibility on the perpetrator. We present a quasi-parallel dataset of sentences with low and high perceived responsibility levels for the perpetrator, and experiment with unsupervised (mBART-based), zero-shot and few-shot (GPT3-based) methods for rewriting sentences. We evaluate our models using a questionnaire study and a suite of automatic metrics.
\end{abstract}

\section{Introduction}
\label{sec:introduction}

``A terrible incident involving husband and wife'', ``Husband kills wife", ``Her love for him became fatal'': these different phrasings can all be used to describe the same violent event, in this case a \textit{femicide}, but they won't trigger the same perceptions in the reader. Perceptions vary from person to person, of course, but also depend substantially and systematically on the different ways the same event is framed \citep{iyengar1994anyone}. Especially in the context of gender-based violence (GBV), this has important consequences on how readers will attribute responsibility: victims of femicides are often depicted, and thus perceived, as (co-)responsible for the violence they suffer.\footnote{A report on femicides from November 2018 by two Italian research institutes points out that the stereotype of a shared responsibility between the victim and its perpetrator is still widespread among young generations: “56.8\% of boys and 38.8\% of girls believe that the female is at least partly responsible for the violence she has suffered" (Laboratorio Adolescenza and Istituto IARD, 2018).
}

\begin{textbox}
    \centering
    \tcbset{colback=white,colframe=blue!75!black,
    fonttitle=\bfseries}
    \begin{tcolorbox}[enhanced,title=Responsibility Perspective Transfer,
    frame style={left color=red!75!black,right color=blue!75!black}
    % title style image=blueshade.png
    ]
        \small
        \textbf{Definition:} given a sentence $S$ that references an act of violence, write a sentence $S'$ that describes the same facts as $S$ but increases the perceived level of responsibility on the perpetrator of the violence. 

        \vspace{1em}
        \textbf{Examples:} \textit{``A fatal stabbing''} $\mapsto$ \textit{``Someone stabbed another person to death''}\\ \textit{``Woman murdered by husband''} $\mapsto$ \textit{``Man murders wife''}  
        \end{tcolorbox}
    \caption{Task definition}
    \label{box:my_label}
\end{textbox}

%This is a \textbf{tcolorbox}.
%\tcblower
%This is the lower part.
%\tcblower
%further down

There is indeed evidence from the linguistic literature \citep{pinelli2021gender,meluzzi2021responsibility} that people perceive responsibility differently according to how femicides are reported (more blame on the perpetrator in ``Husband kills wife'', more focus on the victim in ``Her love for him became fatal''). In general, linguistic strategies that background perpetrators have been shown to favour victim blaming~\cite{huttenlocher1968,henley1995syntax,bohner2002,gray-wegner2009,hart2020objectification,zhou-etal-2021-assessing}.
This way of reporting contributes to reinforcing such social stereotypes. 

%% should we add some more about backgrounding in general? as we don't have related work section 
%% the below comes from AACL paper.
%%the linguistic backgrounding of agents hinders
%their responsibility and promotes victim blaming~\cite{huttenlocher1968,henley1995syntax,bohner2002,gray-wegner2009,hart2020objectification,zhou-etal-2021-assessing}.

If we want social stereotypes to be challenged, the language we use to describe GBV is thus an excellent place to start, also from a Natural Language Processing (NLP) perspective. Recent work has shown that perspectives on femicides and their triggered perceptions can be modelled automatically \citep{minnema-etal-2022-sociofillmore,minnema-etal-2022-dead}. In this paper, as shown in Box~\ref{box:my_label}, we explore the challenge of \textit{rewriting} descriptions of GBV with the aim to increase the perceived level of blame on the perpetrator, casting it as a style transfer task \cite{xu-etal-2012-paraphrasing, jin-etal-2022-deep}. In this novel \textit{responsibility perspective transfer} task, a given sentence from femicide news reports gets rewritten in a way that puts more responsibility on the perpetrator, while preserving the original content. 

\paragraph{Contributions}
We create an evaluation set containing semi-aligned pairs with ``low'' and ``high'' sentences expressing similar information relative to an event, from an existing dataset of Italian news (\S\ref{sub:datasets}). %RAI femicide data. 
In absence of parallel training data, we follow previous work~\citep{lample2019multipleattribute, fuli-2019,lai-etal-2021-generic} to train an unsupervised style transfer model using mBART~\citep{liu-etal-2020-multilingual-denoising} on non-parallel data (with style labels) with a zero-shot and a few-shot approach using 
%the state-of-the-art large language model 
GPT-3 \citep{brown-gpt3} to perform rewriting (\S\ref{sub:models}). We run both human-based and automatic evaluations to assess the impact of rewriting on the perceived blame, comparing original and rephrased texts to find that models can achieve detectable perspective shifts (\S\ref{sec:results}). By introducing the novel task of responsibility perspective transfer, providing an evaluation dataset, a battery of trained models, and evidence of a successful methodology, we hope to foster further research and application developments on this and other perspective rewriting tasks that are relevant to society.\footnote{Data and code are available at \url{https://github.com/gossminn/responsibility-perspective-transfer}.}

% \begin{itemize}
%     \item \textbf{Problem:} writing and perception; common framing. 
%     \item \textbf{Task definition:} rewrite sentence from femicide news reports in order to increase the perceived level of blame on the perpetrator
%     \item \textbf{Approach:} start from semi-aligned dataset based on RAI corpus, explore different semi-supervised and few-shot methods (notably, based on GPT-3)
% \end{itemize}

% \section{Related work}
% \label{sec:related-work}

% \subsection{Responsibility Perception}
% ...

%\subsection{Text Style Transfer}

%In this work, we explore the problem of rewriting descriptions of gender-based violence (GBV), and introduce a novel style transfer task based on GBV descriptions as a means of changing the perceived level of blame on the perpetrator.

% \todo{MN: I would swap the order of subsecs\\Huiyuan: Yeah, although we might not need a 'related work' Sec for a short paper, but let's see..}

\section{Experimental Settings}
\label{sec:methods}

% \subsection{Responsibility Perspective Transfer}
% \gossecamready{The goal of textual style transfer is to alter the style of given texts while maintaining their original content (i.e., the main meaning). Here, style is the perceived level of responsibility, divided into low and high levels. For example, ...
% Formally, given a sentence $x$ with a low level of blame, model $M$ can rewrite it as $y$ with a high perceived level of blame on the perpetrator while preserving the original content: $M (x, blame_{low}) = y$. \todo{check}}

\begin{table*}[t]
\centering
\resizebox{.9\textwidth}{!}{%
\begin{tabular}{@{}lc|cc|ccc|cccc@{}}
\toprule
\multicolumn{2}{c|}{\textbf{Perspective Model}} &
  \multicolumn{1}{l}{\textbf{Source}} &
  \multicolumn{1}{l|}{\textbf{Target}} &
  \multicolumn{3}{c|}{\textbf{mBART}} &
  \multicolumn{4}{c}{\textbf{GPT-3}} \\
\multicolumn{1}{c}{\textit{\textbf{Dimension}}} &
  \textit{\textbf{R$^2$}} &
  \textbf{} &
  \textit{\textbf{(avg)}} &
  \textit{\textbf{base}} &
  \textit{\textbf{src-meta}} &
  \textit{\textbf{meta-src}} &
  \textit{\textbf{na-zero}} &
  \textit{\textbf{na-few}} &
  \textit{\textbf{iter-1}} &
  \textit{\textbf{iter-2}} \\ \midrule
{\color[HTML]{00009B} \textbf{"blames the murderer"}} &
  \textit{0.61} &
  \cellcolor[HTML]{F3C0BC}{\color[HTML]{00009B} -0.511} &
  \cellcolor[HTML]{D8F0E4}{\color[HTML]{00009B} 0.445} &
  \cellcolor[HTML]{F9E0DE}{\color[HTML]{00009B} -0.250} &
  \cellcolor[HTML]{FAE8E6}{\color[HTML]{00009B} -0.188} &
  \cellcolor[HTML]{E7F5EE}{\color[HTML]{00009B} 0.284} &
  \cellcolor[HTML]{FBEBEA}{\color[HTML]{00009B} -0.157} &
  \cellcolor[HTML]{F6D1CE}{\color[HTML]{00009B} -0.375} &
  \cellcolor[HTML]{F6FCF9}{\color[HTML]{00009B} 0.109} &
  \cellcolor[HTML]{FCF0EF}{\color[HTML]{00009B} -0.116} \\
{\color[HTML]{000000} \textbf{"caused by a human"}} &
  \textit{0.60} &
  \cellcolor[HTML]{F9E3E1}{\color[HTML]{000000} -0.228} &
  \cellcolor[HTML]{E0F3E9}{\color[HTML]{000000} 0.362} &
  \cellcolor[HTML]{FEFAFA}{\color[HTML]{000000} -0.037} &
  \cellcolor[HTML]{FFFFFF}{\color[HTML]{000000} 0.005} &
  \cellcolor[HTML]{DFF2E9}{\color[HTML]{000000} 0.371} &
  \cellcolor[HTML]{FCFEFD}{\color[HTML]{000000} 0.042} &
  \cellcolor[HTML]{FCF3F2}{\color[HTML]{000000} -0.095} &
  \cellcolor[HTML]{E7F6EE}{\color[HTML]{000000} 0.278} &
  \cellcolor[HTML]{F9FDFB}{\color[HTML]{000000} 0.076} \\
{\color[HTML]{000000} \textbf{"focuses on the murderer"}} &
  \textit{0.65} &
  \cellcolor[HTML]{F3C0BB}{\color[HTML]{000000} -0.518} &
  \cellcolor[HTML]{CBEADB}{\color[HTML]{000000} 0.597} &
  \cellcolor[HTML]{FAE8E7}{\color[HTML]{000000} -0.184} &
  \cellcolor[HTML]{FCF1F1}{\color[HTML]{000000} -0.108} &
  \cellcolor[HTML]{CEEBDD}{\color[HTML]{000000} 0.567} &
  \cellcolor[HTML]{FDFEFE}{\color[HTML]{000000} 0.033} &
  \cellcolor[HTML]{F6D4D1}{\color[HTML]{000000} -0.349} &
  \cellcolor[HTML]{F0F9F5}{\color[HTML]{000000} 0.179} &
  \cellcolor[HTML]{FCF2F1}{\color[HTML]{000000} -0.104} \\
% {\color[HTML]{656565} \textbf{"focuses on the victim"}} &
%   {\color[HTML]{656565} \textit{0.63}} &
%   \cellcolor[HTML]{FEFFFE}{\color[HTML]{656565} 0.019} &
%   \cellcolor[HTML]{FBEEED}{\color[HTML]{656565} -0.137} &
%   \cellcolor[HTML]{FCFEFD}{\color[HTML]{656565} 0.035} &
%   \cellcolor[HTML]{F1F9F5}{\color[HTML]{656565} 0.171} &
%   \cellcolor[HTML]{F0F9F5}{\color[HTML]{656565} 0.173} &
%   \cellcolor[HTML]{FCF0EF}{\color[HTML]{656565} -0.123} &
%   \cellcolor[HTML]{F0F9F5}{\color[HTML]{656565} 0.178} &
%   \cellcolor[HTML]{EEF9F4}{\color[HTML]{656565} 0.195} &
%   \cellcolor[HTML]{F7FCFA}{\color[HTML]{656565} 0.097} \\ 
  % \hline
  % \hline
  % harm. mean\\
\bottomrule
\end{tabular}%
}
\caption{Automatic evaluation of perspective using the BERTino-based model from \citet{minnema-etal-2022-dead}. Scores are z-normalized (i.e., a score -1 or 1 means ``one standard deviation below/above average''). Target scores are averaged across different target sentences.}
\label{tab:auto-perspective}
\end{table*}

\subsection{Datasets}
\label{sub:datasets}

Our work makes use of the \textit{RAI femicide corpus} \cite{belluati2021femminicidio}, a dataset of 2,734 news articles covering 582 confirmed femicide cases and 198 other GBV-related cases\footnote{Including cases of non-lethal violence, suspected femicide, and suicide.} in Italy between 2012-2017. Of these, 182 cases (comprising 178 femicides and 4 other cases) are linked to a set of news articles from the period 2015-2017 that report on these cases. This dataset is augmented with perspective annotations from \citet{minnema-etal-2022-dead}. Gold annotations (averaged z-scored perception values from 240 participants) are available for 400 sentences, and silver annotations (annotated with the best-scoring model from \citealt{minnema-etal-2022-dead}) are available for 7,754 further sentences. Using event metadata, we automatically extracted pairs of sentences $\langle L,H\rangle$, where $L$ and $H$ both reference the same GBV case, but respectively have a below-average ($L$) or above-average ($H$) level of perceived perpetrator blame. Next, for a subset of 1,120 sentences from the combined gold-silver perspective dataset, we performed manual filtering to ensure that for pair, $L$ and $H$ reference not only the same \textit{case}, but also show substantial overlap in terms of the specific \textit{events} within this case that they describe (e.g. the violence itself, the police investigation, conviction of a suspect, etc.). This yielded a set of 2,571 pairs (or 304 pairs if each sentence is used only once). 

\subsection{Models}
\label{sub:models}

Due to the limited availability of parallel data, we experiment with several existing text generation methods known to work in low-data settings.

\paragraph{Unsupervised mBART} We train an unsupervised model with iterative back-translation~\citep{hoang-etal-2018-iterative}: two mBART-based models, one for each transfer direction, where outputs of one direction with source sentences are used to supervise the model in the opposite direction. 
All experiments are implemented atop Transformers~\citep{wolf-etal-2020-transformers} using mBART-50~\citep{tang-etal-2021-multilingual}. We use the Adam optimizer with a polynomial learning rate decay, and a linear warmup of 100 steps for a maximum learning rate of 1e-4. We limit the maximum token length to 150. To alleviate computational costs and catastrophic forgetting, we only update the parameters of the decoder, freezing the other parameters.

% \item
\paragraph{mBART + meta-information} A unique feature of our dataset is the availability of detailed meta-information about the events. We made a selection of the properties likely to be most relevant for characterizing the event and assigning responsibility (names of the victim and perpetrator, type of victim-perpetrator relationship, murder weapon and location)  and concatenated this meta-information to the corresponding source sentence as input during training. We tried two order settings: \textbf{\textit{source-meta}} and \textbf{\textit{\textit{meta-source}}}. Preliminary experiments showed that concatenating only the event properties themselves, without including property names, produced the most promising results. %Appendix~\ref{app:examples} shows an example.  
For example: \textit{``Trapani, Donna di 60 anni uccisa dall'ex marito --- Anna Manuguerra, Antonino Madone, ex coniuge, arma da taglio, Nubio, casa'' } (``Trapani: 60-year old woman killed by ex-husband --- [victim name], [perpetrator name], ex-spouse, cutting weapon, [town name], at home''). We use the same training setup as for the previous model.

% \item 
\paragraph{GPT-3: Naive implementation} We also experimented with using the \textit{text-davinci-002} version of GPT-3~\cite{brown-gpt3} in a range of zero-shot and few-shot setups. Our \textbf{\textit{naive-zero}} setup uses a simple prompt telling the model to rewrite the sentence with more focus on the perpetrator.\footnote{\textit{``Riscrivi la frase concentrandoti sul colpevole''} (``Rewrite the sentence and concentrate on the culprit'')} Next, \textbf{\textit{naive-few}} uses a similarly simple prompt\footnote{\textit{``Riscrivi le seguenti frasi    da low ad high. Per high si intende che la colpa è attribuita interamente al killer. Ecco alcuni esempi: [...] Riscrivi la seguente frase:''} (``Rewrite the following sentences from low to high. `High' means that the blame is entirely put on the killer. Here are some examples: [...] Rewrite the following sentence:'')} along with a set of ten low-high sentence pairs randomly sampled from the gold annotations.  

% \item
\paragraph{GPT-3: Iterative few-shot} A challenging factor for our naive few-shot approach is that the the `natural' source-target pairs from our annotated data are not perfect minimal pairs, as they differ in perspective but also have some content differences. In an effort to use maximally informative pairs as few-shot examples, we designed an iterative process for compiling small curated sets of examples. First, we designed an improved zero-shot prompt by giving a set of source-target pairs sampled from the gold annotations to the model and prompting it to explain the differences between the pairs. We discovered by accident that this yields a very plausible and concise task definition, and we reasoned that a definition generated by the model on the basis of real examples might be more informative as a prompt than a manually designed one. We then provided two annotators\footnote{The annotators were authors G.M. and M.B.} with the resulting definition\footnote{The definition (slightly edited for grammar) is: \textit{``Le frasi precedute dall'etichetta "Low:" tendono ad essere più brevi e non danno la colpa esplicita all'assassino, mentre le frasi precedute dall'etichetta "High:" tendono ad essere più dirette e a dare la colpa all'assassino.''} (``The sentences preceded by "Low:" tend to be shorter and don't explicitly blame the murderer, while the sentences preceded by "High:" tend to be more direct and blame the murderer.'')}, as well as with five more source sentences sampled from the corpus. Each of the annotators then adapted the definition into a zero-shot prompt, used that prompt to generate target sentences for each of the source sentences, and selected the best candidate from these to create a set of pairs with maximal perspective contrast and content overlap, to be used in a few-shot prompt. We kept both versions of the few-shot prompt, \textbf{\textit{iter-1}} and \textit{\textbf{iter-2}} in order to measure the combined effects of small difference in prompt, randomness in the generated candidates, and judgement differences in the selection of the best candidate.

\begin{table*}[t]
\centering
\resizebox{.8\textwidth}{!}{%
\begin{tabular}{@{}lc|ll|ccc|cccc@{}}
\toprule
\textbf{Metric} &
  \multicolumn{1}{l|}{\textbf{$\leftrightarrow$}} &
  \textbf{Source} &
  \textbf{Target} &
  \multicolumn{3}{c|}{\textbf{mBART}} &
  \multicolumn{4}{c}{\textbf{GPT3}} \\
 &
  \multicolumn{1}{l|}{} &
   &
  \textit{\textbf{(avg)}} &
  \multicolumn{1}{l}{\textit{\textbf{base}}} &
  \multicolumn{1}{l}{\textit{\textbf{src-meta}}} &
  \multicolumn{1}{l|}{\textit{\textbf{meta-src}}} &
  \multicolumn{1}{l}{\textit{\textbf{na-zero}}} &
  \multicolumn{1}{l}{\textit{\textbf{na-few}}} &
  \multicolumn{1}{l}{\textit{\textbf{iter-1}}} &
  \multicolumn{1}{l}{\textit{\textbf{iter-2}}} \\ \midrule
\textbf{BLEU} &
  src &
  \multicolumn{1}{c}{-} &
  \multicolumn{1}{c|}{\cellcolor[HTML]{CDE1F3}0.015} &
  \cellcolor[HTML]{669FD3}0.725 &
  \cellcolor[HTML]{76AAD8}0.612 &
  \cellcolor[HTML]{ADCDE9}0.236 &
  \cellcolor[HTML]{A3C6E6}0.303 &
  \cellcolor[HTML]{90BAE0}0.435 &
  \cellcolor[HTML]{88B5DD}0.489 &
  \cellcolor[HTML]{A6C8E7}0.285 \\
\textbf{ROUGE} &
  src &
  \multicolumn{1}{c}{-} &
  \multicolumn{1}{c|}{\cellcolor[HTML]{F2C2C2}0.100} &
  \cellcolor[HTML]{E47A7A}0.808 &
  \cellcolor[HTML]{E68585}0.701 &
  \cellcolor[HTML]{EDA9A9}0.351 &
  \cellcolor[HTML]{E99494}0.551 &
  \cellcolor[HTML]{E88B8B}0.638 &
  \cellcolor[HTML]{E78989}0.659 &
  \cellcolor[HTML]{EB9F9F}0.450 \\
\textbf{COMET} &
  src &
  \multicolumn{1}{c}{-} &
  \multicolumn{1}{c|}{\cellcolor[HTML]{CBC2E2}-1.216} &
  \cellcolor[HTML]{AA9CD1}0.540 &
  \cellcolor[HTML]{AFA2D4}0.257 &
  \cellcolor[HTML]{BFB4DC}-0.591 &
  \cellcolor[HTML]{B2A5D6}0.103 &
  \cellcolor[HTML]{AA9CD1}0.538 &
  \cellcolor[HTML]{AD9FD3}0.379 &
  \cellcolor[HTML]{B5A9D7}-0.058 \\
\textbf{BLEU} &
  tgt &
  \multicolumn{1}{c}{\cellcolor[HTML]{CDE1F3}0.015} &
  \multicolumn{1}{c}{-} &
  \cellcolor[HTML]{CDE1F3}0.014 &
  \cellcolor[HTML]{CDE1F3}0.016 &
  \cellcolor[HTML]{CCE0F2}0.024 &
  \cellcolor[HTML]{CEE2F3}0.010 &
  \cellcolor[HTML]{CEE1F3}0.013 &
  \cellcolor[HTML]{CDE1F3}0.014 &
  \cellcolor[HTML]{CEE2F3}0.009 \\
\textbf{ROUGE} &
  tgt &
  \multicolumn{1}{c}{\cellcolor[HTML]{F2C2C2}0.100} &
  \multicolumn{1}{c}{-} &
  \cellcolor[HTML]{F2C1C1}0.110 &
  \cellcolor[HTML]{F2C2C2}0.104 &
  \cellcolor[HTML]{F2BFBF}0.132 &
  \cellcolor[HTML]{F3C4C4}0.088 &
  \cellcolor[HTML]{F3C3C3}0.094 &
  \cellcolor[HTML]{F3C3C3}0.098 &
  \cellcolor[HTML]{F3C3C3}0.090 \\
\textbf{COMET} &
  tgt &
  \multicolumn{1}{c}{\cellcolor[HTML]{CBC2E2}-1.175} &
  \multicolumn{1}{c}{-} &
  \cellcolor[HTML]{CAC1E2}-1.194 &
  \cellcolor[HTML]{CAC1E2}-1.178 &
  \cellcolor[HTML]{C7BDE0}-1.002 &
  \cellcolor[HTML]{C8BFE1}-1.090 &
  \cellcolor[HTML]{C8BEE0}-1.045 &
  \cellcolor[HTML]{C8BEE1}-1.057 &
  \cellcolor[HTML]{C8BEE1}-1.059 \\ \bottomrule
\end{tabular}%}
}
\caption{Automatic content preservation metrics (BLEU, ROUGE, COMET), comparing generated sentences against source and gold target sentences.}
\label{tab:bleu-rouge-comet}
\end{table*}
\begin{table}[t]
\centering
\resizebox{.4\textwidth}{!}{%
\begin{tabular}{@{}ll|ccc@{}}
\toprule
\multicolumn{2}{c}{\textbf{}} &
  \multicolumn{1}{c}{\textbf{Perspective}} &
  \multicolumn{1}{c}{\textbf{Similarity}} &
  \multicolumn{1}{c}{\textbf{HM}} \\ \midrule
  \textbf{mBART} &
  \multicolumn{1}{c}{\textit{base}} &
  \cellcolor[HTML]{E2BFD0}2.14 &
  \cellcolor[HTML]{8BB3BB}7.72 &
  \cellcolor[HTML]{D7EAD1}3.34 \\
  & \multicolumn{1}{c}{\textit{src-meta}} &
  \cellcolor[HTML]{E0BCCD}2.50 &
  \cellcolor[HTML]{93B8C0}6.78 &
  \cellcolor[HTML]{BDECB8}3.65 \\
  & \multicolumn{1}{c}{\textit{meta-src}} &
  \cellcolor[HTML]{D8ABC1}4.50 &
  \cellcolor[HTML]{B0CBD1}3.62 &
  \cellcolor[HTML]{A0EF9C}4.01 \\ \hline
  \textbf{GPT-3} &
  \multicolumn{1}{c}{\textit{na-zero}} &
  \cellcolor[HTML]{DFBACC}2.77 &
  \cellcolor[HTML]{96BAC2}6.52 &
  \cellcolor[HTML]{AAEEA5}3.89 \\
  & \multicolumn{1}{c}{\textit{na-few}} &
  \cellcolor[HTML]{E2C0D0}2.08 &
  \cellcolor[HTML]{87B0B9}8.17 &
  \cellcolor[HTML]{D9EAD3}3.31 \\
  & \multicolumn{1}{c}{\textit{iter-1}} &
  \cellcolor[HTML]{DCB2C7}3.57 &
  \cellcolor[HTML]{89B1BA}7.97 &
  \cellcolor[HTML]{50F74E}4.98 \\
  & \multicolumn{1}{c}{\textit{iter-2}} &
  \cellcolor[HTML]{DBB0C5}3.84 &
  \cellcolor[HTML]{95BAC1}6.60 &
  \cellcolor[HTML]{5AF658}4.85 \\ \hline
  \textbf{Examples} & 
  \multicolumn{1}{c}{\textit{for iter-1}} &
  \cellcolor[HTML]{D6A5BD}5.20 &
  \cellcolor[HTML]{92B8BF}6.93 &
  \cellcolor[HTML]{00FF00}5.94 \\
  & \multicolumn{1}{c}{\textit{for iter-2}} &
  \cellcolor[HTML]{DBB0C5}3.87 &
  \cellcolor[HTML]{A1C1C8}5.27 &
  \cellcolor[HTML]{7BF377}4.46 \\ 
  \bottomrule
\end{tabular}%
}\caption{Human evaluation results on model outputs and examples for few-shot. HM is the harmonic mean of perspective and similarity scores}
\label{tab:hum}
\end{table}

\subsection{Evaluation Methods}

The main goal of responsibility perspective transfer is to generate a sentence with the desired perspective (``style strength" in classic style transfer tasks) that still has the same semantic content as the source sentence. We assess the performance of different models using standard metrics commonly employed in text style transfer~\citep{mir-etal-2019-evaluating, briakou-etal-2021-evaluating, lai-etal-2022-human, jin-etal-2022-deep}, and custom automatic metrics; we also run a questionnaire study with human participants. 

% \gossecamready{In language generation tasks in general, human evaluation is always considered very valuable, but it is not always affordable, especially during development. Therefore, automatic metrics are often more practical and we indeed also used those (both standard metrics such as BLEU and a custom perception-detecting metric taken from previous related work). The perception detectors that we used are derived from a large-scale human study. However, we wanted to have at least a small-scale study with human subjects to assess the validation of the automatic metrics for our new generation task.}

%Regarding the evaluation of this task, most works usually focus on three dimensions: style strength, content preservation, and fluency, which are considered in both automatic and human evaluations~\citep{mir-etal-2019-evaluating, briakou-etal-2021-evaluating, lai-etal-2022-human, jin-etal-2022-deep}.

\paragraph{Automatic Evaluation}

For estimating perspective quality, we used the best-performing perspective regressor from~\citet{minnema-etal-2022-dead} which is based on an Italian monolingual DistilBERT model (\textit{BERTino}; \citealp{Muffo2020BERTinoAI}).

For content preservation, we use three popular text generation metrics: $n$-gram-based \textit{BLEU}~\citep{kishorebleu2002} and \textit{ROUGE}~\citep{lin-2004-rouge}, as well as a neural-based model \textit{COMET}~\citep{rei-etal-2020-comet}.

\paragraph{Human Evaluation} Participants were given an online survey with 50 blocks, each corresponding to one source sentence sampled from the dataset. In each block, participants rated: 1) the level of perceived agent responsibility in each of the seven \textit{target} candidates; 2) the level of \textit{content preservation} of each target relative to the source. We also designed a separate, smaller questionnaire that asked the same questions about the few-shot examples used in \textbf{\textit{iter-1}} and \textbf{\textit{iter-2}}. 

The pool of invited participants was a group of people with mixed genders and backgrounds from the personal network of the authors. No remuneration was offered. Four invitees responded to the main questionnaire, and three invitees responded to the few-shot example questionnaire (all female, mean age: 46). The participants have different levels of education (from middle school to university) and live in different regions of Italy. 

Our evaluation study should be seen as a pilot, and larger-scale, more representative studies are planned for the future. The main aim of the pilot was to have a small-scale validation of our automatic metrics (taken from previous work and developed on the basis of a large-scale human study) and to test our evaluation setup (which questions to ask, etc.).
% Participants (n=4 for main questionnaire; n=3 for the few-shot examples; all female; mean age 46) were native Italian speakers recruited from the authors' personal network; no remuneration was offered. 
The questionnaire was designed %(automatically generated with custom scripts) 
and distributed using Qualtrics.\footnote{\url{https://www.qualtrics.com/}}

\section{Results}
\label{sec:results}

\subsection{Automatic Results}

\paragraph{Perspective Evaluation} Following \citet{minnema-etal-2022-dead}, we distinguish between several perceptual dimensions using a perception regression model, as shown in Table~\ref{tab:auto-perspective}. Our main dimension of interest (highlighted in blue) is \textit{blame on murderer}, but we also look at the two closely related dimensions of \textit{cause} and \textit{focus on murderer}. As shown by the $R^2$ scores, regression quality is decent for all of these dimensions. We observe that the source and target sentences have lower and higher blame scores respectively, which are also consistent on the two related dimensions, affirming that our testing data is of good quality in terms perspective aspect.

For all models, the perception scores of the predicted sentences are higher than those of the source sentences, with mBART/\textit{meta-src} achieving the highest scores. This suggests that all models alter perceptions of responsibility to some extent. However, in virtually all cases, perception scores stay well below the target, and in many cases below the average level (zero).  For mBART-based results, models with meta-information perform better than the baseline, with \textit{meta-src} reaching particularly high scores. Within the GPT-3 settings, zero-shot (\textit{na-zero}), surprisingly, performs better than few-shot; (\textit{na-few}), and \textit{iter-1} yields the highest scores. 

% As a final note, we find that perception on the \textit{focus on victim} dimension changes in the opposite direction: while the target sentences score lower than the source sentences, most of the predictions end up with a higher value than the source. 

\paragraph{Content Preservation}

When taking source sentences as the reference, three metrics show that the outputs have higher similarities to them than the target sentences. mBART/\textit{base} has the highest scores, which (combined with the low perception scores of this model) suggests that the model tends to copy from the source sentence. Within the GPT-3 settings, \textit{iter-1} has the highest scores. Using instead the target sentences as reference, we see that all scores are very close, with mBART/\textit{meta-src}) reaching the best performance, followed by GPT-3/\textit{na-few} and GPT-3/\textit{iter-1}.

\subsection{Human-based Results}

% %\paragraph{Inter-annotator agreement}

% \todo{add sentence on agreement}
%\paragraph{Results} 
Table~\ref{tab:hum} reports the results of our human evaluation study. We find that mBART/\textit{meta-src} is the best overall model on perspective, but has poor similarity. Meanwhile, GPT3/\textit{na-few} achieves the highest score on similarity but the lowest score in terms of perspective, and its overall performance is lower than that of GPT3/\textit{na-zero}. GPT3/\textit{iter-1} has the best overall performance with an HM of 4.98. We found reasonably high levels of inter-annotator agreement (Spearman's rank correlation between pairs of annotators). Correlations ranged between 0.3-0.6 (blame) and 0.4-0.6 (similarity) with high levels of significance ($p<0.0001$). The examples for few-shot are of higher quality overall as they were picked by the authors. 
%It is also interesting to see that better examples lead to more successful outputs.

\subsection{Case Study}

Box~\ref{box:examples} shows two sets of example outputs generated by mBART and GPT-3.\footnote{Due to lack of space, we include only generations from the overall best-performing model from each category.} While hand-picked, these examples show that both models are capable of generating sentences that increase responsibility while trying to preserve content. However, they also highlight a key challenge: what if the source sentence lacks details about the event? The mBART model has access to event metadata and uses this effectively in Example~1 to produce a sentence that stays close to the source but with details from the metadata filled in (though with rather clunky sentence structure). In Example~2, instead, it produces a sentence that is factually correct but also loses most of the information from the source sentence. On the other hand, GPT-3, without access to metadata, often `invents' missing information. This is evident in the second example, in which it faithfully preserves the source sentence and increases the level of blame by adding plausible but (partially) incorrect information about the crime.

\begin{textbox}[t]
    \centering
        \tcbset{colback=white,colframe=blue!75!black,
        fonttitle=\bfseries}
        \begin{tcolorbox}[enhanced,title=Prediction Examples,
            frame style={left color=red!75!black,right color=blue!75!black}
        % title style image=blueshade.png
        ]
            \small
            
            \textbf{(Ex. 1)} \textit{Provaglio d'Iseo, donna trovata morta in casa : si sospetta il compagno} ``[town name], woman found dead at home: [her] partner is suspected''

            \vspace{.5em}
            \textcolor{RoyalPurple}{\textbf{Target:} \textit{Donna massacrata in casa nel Bresciano, il convivente fermato per omicidio volontario} ``Woman massacred at home in [area name]: [her] cohabitant [partner] arrested for voluntary homocide''}

            \vspace{.5em}
            \textcolor{Bittersweet}{\textbf{mBART (meta-src):} \textit{Un uomo, Elio Cadei, è stato fermato per l'omicidio di Simona Simonini, la donna trovata morta in casa: si sospetta il compagno di lei.} ``A man, Elio Cadei, has been arrested for the murder of Simona Simonini, the woman found dead at home: the partner of her is suspected''}
            
            \vspace{.5em}
            \textcolor{Bittersweet}{\textbf{GPT-3 (iter-1):} \textit{Provaglio d'Iseo, donna trovata morta in casa: il compagno è il principale sospettato} ``[town name], woman found dead at home:    [her] partner is the principal suspect''}
            
            \vspace{.5em}
            \hrule
            \vspace{.5em}
            \textbf{(Ex. 2)} \textit{«Come vuole che ci si senta, quando ci sono stati due morti?»} ``«How do you want us to feel, when there have been two deaths?»''

            \vspace{.5em}
            \textcolor{RoyalPurple}{\textbf{Target:} \textit{Albenga, uccide l'ex compagna a coltellate e poi si toglie la vita Loredana e l'incubo-stalker foto} ``[town name], [someone] kills [his] ex-partner by stabbing and then takes his life[.] Loredana and the nightmare stalker[, see] foto''}

            \vspace{.5em}
            \textcolor{Bittersweet}{\textbf{mBART (meta-src):} \textit{Un uomo, Mohamed Aziz El Mountassir, di 43 anni, ha ucciso la moglie, Loredana Colucci, di 44, con una coltellata al collo.} ``A man, Mohamed Aziz El Mountassir, 43 years old, has killed his wife, Loredana Colucci, 44 years old, with a knife to [her] neck.''}

            \vspace{.5em}
            \textcolor{Bittersweet}{\textbf{GPT-3 (iter-1):} \textit{«Come vuole che ci si senta, quando ci sono stati due morti causati da un uomo che ha ucciso prima la sua ex moglie e poi la sua nuova compagna?} ``How do you want us to feel, when there have been two deaths caused by a man who has first killed his ex-wife and then his new partner?''} 
            
        \end{tcolorbox}
    \caption{Prediction examples}\label{box:examples}
\end{textbox}

\section{Discussion \& Conclusion}
\label{sec:discussion}

We proposed responsibility perspective transfer as a new task and introduced a dataset and models for applying this task to Italian news reporting about femicides. Our dataset contains a limited amount of quasi-aligned pairs that proved useful for evaluation and few-shot learning. We experimented with two modeling approaches: unsupervised mBART (with or without enriching the input with metadata) and zero-shot/few-shot learning with GPT-3. 

Our human and automatic evaluations suggest \textit{GPT-3/iter-1} as the best overall model, with a relatively high level of responsibility placed on the perpetrator and a good degree of content preservation. For the latter, %Content preservation seems to have been achieved quite well, with 
most models score at least 6/10 on average on the human survey. The perspective change itself has also been achieved by our models, with substantially increased levels of perceived perpetrator blame compared to the source, but there is still much room for improvement: none of the models comes close to having the same level of blame as the target sentences do, and in the human evaluation survey no model achieves a `blame score' of more than 4.5/10. The main obstacle for future improvements seems to lie with the lack of truly parallel data; however, our GPT-3-based iterative approach of creating minimal pairs 
seems to have worked quite well, and might be further exploited on a  larger scale.

% \clearpage

\section{Limitations}
This paper introduced the new task of responsibility perspective transfer and provided initial data collection and modeling for a specific domain (news about gender-based violence) and language (Italian). The main limitation of our work is that the (mBART) models that we trained and the prompts (for GPT-3) that we designed are specific to this domain and language and cannot be applied `out-of-the-box' in other contexts. However, all of our modeling setups require no or limited training data and make use of readily available existing models, so we believe the general approach to be easily transferrable to other domains. 

Another limitation comes from the fact that we used GPT-3: the model is closed-source and can only be accessed with a paid subscription to the OpenAI API (\url{https://beta.openai.com/}). This has consequences for reproducibility for several reasons. First of all, we do not have access to the exact technical specifications of the model or to the training data that was used. The GPT-3 models are regularly updated (at the time of our experiments, \textit{text-davinci-002} was the most recent available version), but limited information is available about what distinguishes each version from the previous ones or from the original model introduced in \citet{brown-gpt3}. Moreover, access to the API is controlled by OpenAI and could be closed at any time at the company's discretion; the API is currently quite accessible with no waiting list and a reasonably generous free trial, but the rates (paid in USD) might not be affordable for researchers outside of institutions in high-income countries, and not all researchers might be comfortable agreeing to the company's terms and conditions. Finally, the generative process involves a degree of randomness, and through the API it is not possible to fixate the model's random seed, meaning that the model produces different predictions every time it is called, even when using exactly the same prompt.  

\section{Ethics Statement}

We see three important ethical considerations around our paper. The first consideration is related to the use of large proprietory language models (GPT-3). Apart from the reproducibility limitations resulting from the use of GPT-3 discussed above, there are more general ethical questions surrounding the use of GPT-3 and similar models, for example the high energy usage and resulting carbon emissions, and societal questions around the oligopoly on state-of-the-art language models that is currently in the hands of a handful of large US-based companies.

The second consideration relates to the task that we introduce: while we see perspective transfer models as a valuable tool for studying how language `frames' (social) reality that could also have practical applications, for example in journalism, we strongly believe that any such applications must be approached with extreme care. The models that we introduce are scientific analysis tools that could be used to suggest alternative viewpoints on an event, but we believe that generations should \textit{not} be seen as necessarily reflecting a `true' or `better' perspective, and should not used in a prescriptive way (i.e. used to tell someone how to write). We believe that the authors (journalists or others) of any text ultimately bear exclusive responsibility for the views, perspectives and (implicit) values expressed in it, and should be careful in making use of texts (re-)written by computers, such as the ones produced by our proposed models.

Finally, we are aware that our task domain (femicide/gender-based violence) is a societally and emotionally loaded topic, and that the texts contained in our dataset and produced by our models might be disturbing. In particular, in some cases, models may produce graphic descriptions of violence and/or produce questionable moral judgements (e.g., we have occasionally seen statements such as ``the perpetrator of this horrible crime does not have the right to live'' spontaneously produced by some of the models), and potential users of applications of the model should be aware of this. For the purposes of this paper, the only people external to the research team who have been extensively exposed to model outputs were the annotators in our human evaluation study. In the introduction page of our online questionnaire, annotators were warned about the sensitive nature of the topic and advised that they could stop their participation at any time if they felt uncomfortable and could contact the authors with any questions. Prior to running the  online questionnaire we have requested and obtained ethical approval by the Ethical Review Committee of our research institution. 

\section*{Author contributions}
Authors G.M. and H.L. share first co-authorship (marked with `*'). G.M. had primary responsibility for data collection and preparation, setting up the GPT-3 experiments and running the human evaluation survey. H.L. had primary responsibility for the mBART experiments and the automatic evaluation. B.M. annotated data (pair alignment) and contributed to prompt engineering and the design of the evaluation questionnaire. M.N. coordinated and supervised the overall project.

\section*{Acknowledgements}
Authors G.M. and M.N. were supported by the Dutch National Science organisation (NWO) through the project \textit{Framing situations in the Dutch language}, VC.GW17.083/6215. Author H.L. was supported by the China Scholarship Council (CSC).

We would like to thank the annotators for helping us evaluate the models' outputs. We also thank the ACL anonymous reviewers for their useful comments. Finally, we thank the Center for Information Technology of the University of Groningen for their support and for providing access to the Peregrine high performance computing cluster. 
\bibliography{anthology,custom}
\bibliographystyle{acl_natbib}

\clearpage
\appendix
\onecolumn
\renewcommand\thefigure{\thesection.\arabic{figure}}
\renewcommand\thetable{\thesection.\arabic{table}}
\setcounter{figure}{0}
\setcounter{table}{0}

\section{Annotation Statistics}
\label{sec:appendix}

\subsection{Inter-annotator agreement}

\begin{figure}[h]
    \begin{minipage}[t]{0.5\linewidth}
    \centering
    \subfigure[Inter-annotator agreement for blame scores in the main human evaluation study.]{
      \includegraphics[scale=0.55]{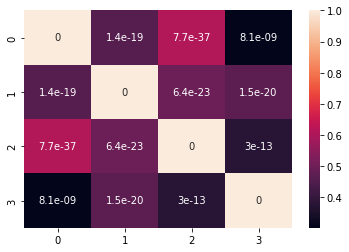}
      \label{fig:app-agree-persp-group1}
    }
    \end{minipage}
    \vspace{1.0mm}
    \begin{minipage}[t]{0.5\linewidth}
    \centering
    \subfigure[Inter-annotator agreement for content preservation scores in the main human evaluation study.]{
      \includegraphics[scale=0.55]{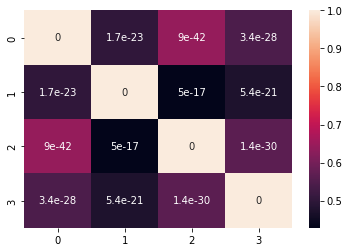}
      \label{fig:app-agr-persp-group1}
    }
    \end{minipage}

    \vspace{1.0mm}
    \begin{minipage}[t]{0.5\linewidth}
    \centering
    \subfigure[Inter-annotator agreement for blame scores in the few-shot prompt evaluation study.]{
      \includegraphics[scale=0.55]{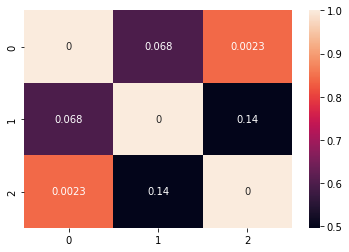}
      \label{fig:app-agr-persp-group1}
    }
    \end{minipage}
    \vspace{1.0mm}
    \begin{minipage}[t]{0.5\linewidth}
    \centering
    \subfigure[Inter-annotator agreement for content preservation scores in the few-shot prompt evaluation study.]{
      \includegraphics[scale=0.55]{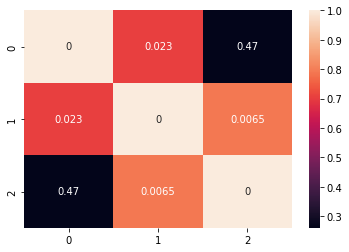}
      \label{fig:app-agr-sim-fewshot}
    }
    \end{minipage}

    \caption{Inter-annotator agreement}
    \label{fig:iaa}
\end{figure}

Figures~\ref{fig:iaa} %through \ref{fig:app-agr-sim-fewshot} 
give inter-annotator agreement scores for the human evaluation. Columns and rows represent individual annotators; colors represent Spearman correlations; numbers in cells are p-values.

% \begin{figure}[h]
%     \centering
%     \includegraphics[scale=.5]{img/human_eval/agreement_perspective_group_1.png}
%     \caption{Inter-annotator agreement for blame scores in the main human evaluation study}
%     \label{fig:app-agree-persp-group1}
% \end{figure}
% \begin{figure}[h!]
%     \centering
%     \includegraphics[scale=.5]{img/human_eval/agreement_similarity_group_1.png}
%     \caption{Inter-annotator agreement for content preservation scores in the main human evaluation study}
%     \label{fig:app-agr-persp-group1}
% \end{figure}

% \begin{figure}[h!]
%     \centering
%     \includegraphics[scale=.5]{img/human_eval/agreement_perspective_fewshot.png}
%     \caption{Inter-annotator agreement for blame scores in the few-shot prompt evaluation study}
%     \label{fig:app-agr-persp-group1}
% \end{figure}

% \begin{figure}[h!]
%     \centering
%     \includegraphics[scale=.5]{img/human_eval/agreement_similarity_fewshot.png}
%     \caption{Inter-annotator agreement for content preservation scores in the few-shot prompt evaluation study}
%     \label{fig:app-agr-sim-fewshot}
% \end{figure}

\clearpage
\section{Questionnaire Materials}

\begin{figure*}[h]
    \centering
    \includegraphics[width=.6\textwidth]{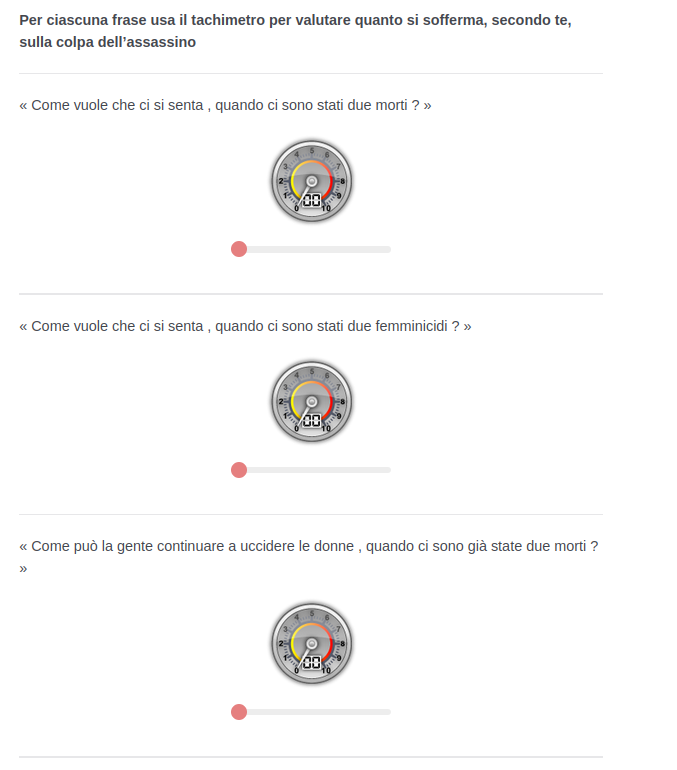}
    \caption{Qualtrics mockup: "speedometer" for rating agentivity}
    \label{fig:qualtrics1}
\end{figure*}

\begin{figure*}[h]
    \centering
    \includegraphics[width=.6\textwidth]{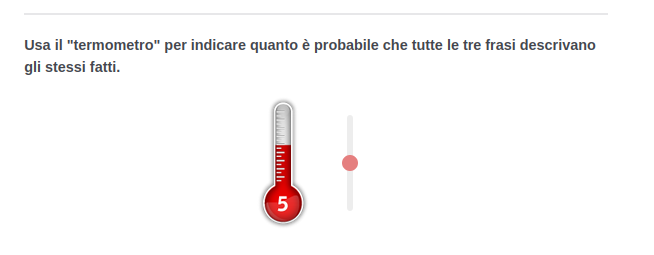}
    \caption{Qualtrics mockup: "thermometer" for rating content preservation}
    \label{fig:qualtrics2}
\end{figure*}

Mockups from the online survey are given in Figures~\ref{fig:qualtrics1} and \ref{fig:qualtrics2}.

% \section{Examples}
% \label{app:examples}

% \paragraph{mBART + meta-information:} \textit{``Trapani, Donna di 60 anni uccisa dall'ex marito --- Anna Manuguerra, Antonino Madone, ex coniuge, arma da taglio, Nubio, casa'' } (``Trapani: 60-year old woman killed by ex-husband --- [victim name], [perpetrator name], ex-spouse, cutting weapon, [town name], at home'').

\end{document}